\documentclass[runningheads]{llncs}

\usepackage{etoolbox}
\usepackage[utf8]{inputenc}
\usepackage[nolist]{acronym}
\usepackage{multirow,array,float,graphicx}
\usepackage{siunitx}
\usepackage[super]{nth}
\usepackage{hyperref}
\usepackage{todonotes}
\usepackage{tabularx}
\usepackage{array}
\usepackage{booktabs}
\usepackage{bm}
\usepackage{amsmath}
\usepackage{amsfonts}
\usepackage[misc]{ifsym}

\hypersetup{
    colorlinks=true,
    pdftitle={Rescoring Sequence-to-Sequence Models for Text Line Recognition with CTC-Prefixes},
    pdfpagemode=FullScreen,
}

\newcounter{rowcntr}[table]
\renewcommand{\therowcntr}{\Alph{rowcntr}}
\newcolumntype{N}{>{\refstepcounter{rowcntr}\therowcntr}c}
\AtBeginEnvironment{tabular}{\setcounter{rowcntr}{0}}

\begin{document}

\title{Rescoring Sequence-to-Sequence Models for Text Line Recognition with CTC-Prefixes}
\titlerunning{Rescoring S2S Models for TLR with CTC-Prefixes}

\author{Christoph Wick\inst{1}\orcidID{0000-0003-3958-6240} \and
Jochen Zöllner \inst{1,2} $^{(\textrm{\Letter})}$\orcidID{0000-0002-3889-6629} \and
Tobias Grüning\inst{1}\orcidID{0000-0003-0031-4942}}
\authorrunning{C. Wick et al.}
\institute{
Planet AI GmbH, \\ Warnowufer 60, 18057 Rostock, Germany \\\email{\{christoph.wick,tobias.gruening\}@planet-ai.de} \and
Computational Intelligence Technology Lab, Department of Mathematics, \\ University of Rostock, 18051 Rostock, Germany \\ \email{jochen.zoellner@uni-rostock.de}
}

\maketitle

\begin{abstract}
    In contrast to \ac{CTC} approaches, \ac{S2S} models for \ac{HTR} suffer from errors such as skipped or repeated words which often occur at the end of a sequence.
    In this paper, to combine the best of both approaches, we propose to use the \ac{CTC}-Prefix-Score during \ac{S2S} decoding.
    Hereby, during beam search, paths that are invalid according to the \ac{CTC} confidence matrix are penalised.
    Our network architecture is composed of a \ac{CNN} as visual backbone, bidirectional \acp{LSTM} as encoder, and a decoder which is a Transformer with inserted mutual attention layers.
    The \ac{CTC} confidences are computed on the encoder while the Transformer is only used for character-wise \ac{S2S} decoding.
    We evaluate this setup on three \ac{HTR} data sets: IAM, Rimes, and StAZH.
    On IAM, we achieve a competitive \ac{CER} of 2.95\% when pretraining our model on synthetic data and including a character-based language model for contemporary English.
    Compared to other state-of-the-art approaches, our model requires about 10-20 times less parameters.
    Access our shared implementations via this  \href{https://github.com/Planet-AI-GmbH/tfaip-hybrid-ctc-s2s}{link to GitHub}.
    
\end{abstract}

\section{Introduction}

\ac{OCR}, the transcription of digital images into machine-actionable text, is still a challenging problem even though there were great advancements in recent years.
A typical \ac{OCR} pipeline consists of two steps: text detection and text recognition.
Currently, most text recognition systems act on lines of text, and thus convert a sequence of arbitrary length (here the width of the text line image) into another sequence of different (typically shorter) length.
There are two established approaches to tackle this problem: \ac{S2S} \cite{sutskever2014sequence} and \ac{CTC} \cite{graves_connectionist_2006}.

\ac{S2S} approaches are a very generic approach as they do not constrain the lengths of the input or output sequence, nor their ordering\footnote{Arbitrary ordering is required, e.g., for translation tasks.}.
An \ac{S2S} system comprises two modules: the \emph{encoder} encodes the line images into features. The \emph{decoder}, starting from a \ac{SOS}-token, decodes the encoded line character-(or token)-wise until an \ac{EOS}-token is emitted.

Since written visual and decoded characters are ordered in the task of \ac{TLR}, the \ac{CTC} algorithm can be applied.
Here, analogous to \ac{S2S}, the line image is first encoded, then \ac{CTC} decodes the complete sequence in one step:
at each position of the encoded feature sequence, \ac{CTC} predicts either the desired character or a special \emph{blank} character, meaning there is no output.
This results in a so-called \emph{confidence matrix} that stores a probability distribution for each character plus the blank for each position in the line.
Best path decoding determines the most likely characters, then repeated character predictions are fused and blanks are erased to obtain the final transcription.

\ac{S2S} has some advantages which are eminent in better accuracies compared to \ac{CTC} (see, e.g., \cite{michael2019evaluating} or \cite{li2021trocr}), but there are also fundamental problems when it comes to \ac{TLR} (see, e.g., \cite{wick2021transformer}):
a crucial problem is that \ac{S2S} approaches tend to predict \ac{EOS}-tokens prematurely or delayed which is why parts of the original sequence are cropped or repeated several times.
Similarly, repetition or skipping of characters, digits, or whole words can also occur within the sequence.
In \ac{TLR}, this can result in severe errors if, for example, repeating digits in a large number are ``skipped''.
The reason for this behaviour is that the mutual-attention modules are not always certain about the snippet that shall be decoded next.
There are attempts to tackle this problem by modifying the attention.
A comparison of six different attention versions, e.g., monotonic (force left to right transcription) or penalised (penalising features that were already attended to), was performed by \cite{michael2019evaluating} for \ac{HTR}.
In contrast, \ac{CTC} approaches do not suffer from any of these problems by design.
Hence, combining the best of both worlds is reasonable but non-trivial since \ac{CTC} and \ac{S2S} have very different concepts for decoding.

Watanabe et al. \cite{watanabe2017hybrid} proposed an algorithm to use \ac{CTC} scores during decoding with an \ac{S2S} approach.
Their network architecture is constructed similar to \ac{S2S} approaches using attention, however, a second \ac{CTC}-based decoder is added to the shared encoder.
Upon decoding, the \ac{CTC}-Prefix-Score is computed to weigh the next character probabilities of the decoder.
Since \cite{watanabe2017hybrid} only applied their approach to speech recognition, in this paper, we examine the applicability on \ac{TLR} which, to the best of our knowledge, has not been performed, yet.
Thereto, we setup a traditional \ac{CNN}/\ac{LSTM} encoder, and a Transformer-based decoder.
Optionally, we add a character-based \ac{LM}.

We evaluate our models on two contemporary and one historic handwritten datasets each in a different language:
IAM \cite{MartiB2002} for English, Rimes \cite{augustin2006Rimes} for French, and StAZH (not published, yet) for historic Swiss-German.

To summarise, in this paper, we contribute the following:
\begin{itemize}
    \item We propose a hybrid CTC/Transformer decoder architecture for \ac{TLR} using both the \ac{CTC} confidence matrix and \ac{S2S} based on \cite{watanabe2017hybrid}.
    \item Additionally, we apply pretraining on synthetic data and a \ac{LM} to achieve state-of-the-art results on IAM.
    \item We share our Tensorflow-based implementation\footnote{\url{https://github.com/Planet-AI-GmbH/tfaip-hybrid-ctc-s2s}} of the network architectures, the \ac{CTC}-Prefix-Scorer, and our beam search.
\end{itemize}

The remainder of this paper is structured as follows:
first, we discuss related work in the area of \ac{TLR} and \ac{HTR}.
Next, we introduce our methodology and describe the computation of \ac{CTC}-Prefix-Scores for joint \ac{CTC}/Transformer-decoding.
Afterwards, we present the three datasets and evaluate our methods.
We conclude with a discussion of our results and future work.

\section{Related Work}

In recent years, \ac{TLR} and \ac{HTR} have been studied quite excessively.
A thorough literature review of handwritten \ac{OCR} is provided by Memon et al. \cite{memon2020handwritten}.

Bluche et al. \cite{bluche2017gated} proposed gated convolutional layers in a \ac{CNN}/\ac{LSTM}/\ac{CTC}-based approach achieving a \ac{CER} of 3.2\% on IAM when using a \ac{LM} with a limited vocabulary size of 50K words.
In \cite{michael2019evaluating}, Michael et al. compare different attention mechanisms in the decoder for \ac{S2S}-based \ac{HTR}.
Their best model on IAM yielded a \ac{CER} of 4.87\% without the usage of external data or a \ac{LM}.

Yousef et al.~\cite{Yousef2020AccurateDU} applied \acp{FCN} without any recurrent connections trained with the \ac{CTC} loss function.
On IAM, they reached a challenging \ac{CER} of 4.9\% without the use of additional data or a \ac{LM}.

Transformers for \ac{TLR} were first introduced by Kang et al. \cite{kang2020pay}.
They proposed a \ac{CNN}/Transformer encoder and a Transformer decoder yielding a \ac{CER} of 4.67\% on IAM when pretraining their model on synthetic data, setting a new best value for open vocabulary \ac{HTR}.
This approach was extended by Wick et al. \cite{wick2021transformer} who proposed a bidirectional decoding scheme: one Transformer reads the line forwards, another one backwards.
The final results is obtained by voting.

Diaz et al.~\cite{diaz2021rethinking} compared different network architectures and decoding schemes for \ac{TLR}.
They varied \acp{RNN}, Gated Recurrent Convolution Layers, and self-attention layers for the encoder, and \ac{CTC} or Trans\-former-based \ac{S2S} for the decoder.
Choosing a \ac{LM} and a pretrained model on an internal real-world dataset, they reached a new state-of-the-art \ac{CER} of 2.75\% using \ac{CTC} and self-attention on the IAM downstream task.

Li et al.~\cite{li2021trocr} proposed a pure \ac{S2S}-Transformer-based approach, i.e., they replaced every \ac{CNN} from the network architecture.
Their large model with 558 million parameters achieved a \ac{CER} of 2.89\% on IAM if pretrained on several millions of synthetic lines.
A separate \ac{LM} was not applied since the decoder was already capable to learn an intrinsic \ac{LM}.

\section{Data}
\label{sec:data}

To evaluate our proposed methods, we use three different datasets for \ac{HTR}: IAM \cite{MartiB2002}, StAZH, and Rimes \cite{augustin2006Rimes}.
Table~\ref{tab:data} summarises the language, alphabet size $\left|A\right|$, and number of lines for training, validation, and testing.

\begin{table}[tbp]
    \centering
    \caption{Overview of the three used datasets showing their language, alphabet size $\left|A\right|$, and the number of training, validation, and test lines.}
    \label{tab:data}
    \begin{tabular}{llr|rrr}
        \toprule
        \bf Dataset & \bf Language & $\bf \left|A\right|$ & \bf \# Train & \bf \# Val & \bf \# Test \\
        \midrule
        IAM & English (en) & 79 & 6,161 & 966 & 2,915 \\
        StAZH & Swiss-German (de-ch) & 109 & 12,628 & 1,624 & 1,650 \\
        Rimes & French (fr) & 100 & 10,171 & 1,162 & 778 \\
        \bottomrule
    \end{tabular}
\end{table}

For the popular IAM-dataset  we use Aachen's partition\footnote{\url{https://github.com/jpuigcerver/Laia/tree/master/egs/iam}}.
StAZH is an internal dataset of the European Union’s Horizon 2020 READ project and thus not openly available, yet.
The documents contain resolutions and enactments of the cabinet as well as the parliament of the canton of Zurich from 1803 to 1882.
We included it for additional results on historical handwritings.

To train our English \ac{LM}, we collected over 16 million text lines from news, webpages, and wikipedia.
For the French \ac{LM}, we used a wikipedia dump with about 30 million text lines.
To train a Swiss-German \ac{LM} only about 1 million text lines of contemporary Swiss-German language were collected.

\section{Methods}

In this section we present our proposed approach for line-based \ac{HTR}.
First, we describe preprocessing steps and our network architecture which comprises the \ac{CTC}- and Transformer-based branches for decoding.
Afterwards, we present the training of the model and how inference including the \ac{CTC}-Prefix-Score and an optional \ac{LM} is performed.
Finally, we describe the training of the character-based \ac{LM} and our line synthesiser for artificial lines used for pretraining the \ac{HTR} models.

\subsection{Preprocessing}
\label{sec:preproc}

To preprocess a line, we apply contrast normalisation without binarisation, normalisation of slant and height, then scale the line images to a fixed height of 64 pixels while maintaining their aspect ratio.
To artificially increase the amount of training data, we augment the preprocessed images by applying minor disturbances to the statistics relevant for the normalisation algorithms.
Furthermore, we combine dilation, erosion, and grid-like distortions \cite{WigingtonSDBPC2017} to simulate naturally occurring variations in handwritten text line images.
These methods are applied to the preprocessed images randomly each with an independent probability of 50\%. 

\subsection{Network-Architecture}

\begin{figure*}[tbp]
    \centering
    \includegraphics[width=\linewidth]{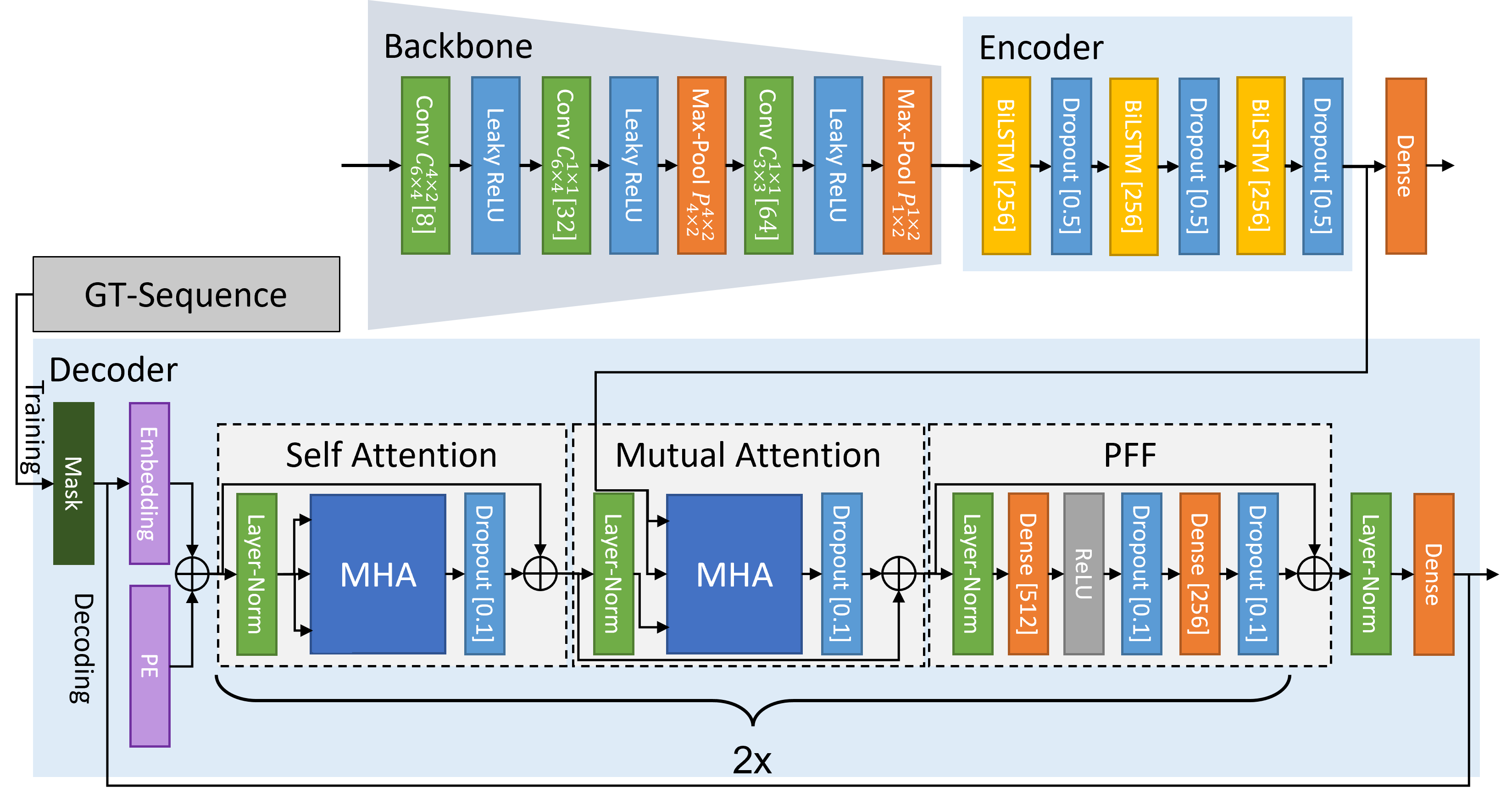}
    \caption{Our proposed network architecture.
    The visual backbone is a \ac{CNN} with a subsampling factor of 8, the encoder comprises three \ac{LSTM}-layers, and the decoder is a Transformer build up by stacked self-attention, mutual-attention layers, and absolute positional encoding (PE).
    }
    \label{fig:network-architecture}
\end{figure*}

Similar to current approaches, we propose a network architecture that is comprised of three basic components (see Figure~\ref{fig:network-architecture}): a visual backbone that extracts high-level features with a limited receptive field, an encoder that can learn context, and a decoder that transcribes the features into characters.
The backbone comprises three convolutional layers and two max-pooling layers which subsample the (rescaled) original line image with a dimension of $64\times W \times 1$ to a feature map of $4\times \frac{W}{8} \times 64$.
The final output is height-concatenated to a dimension of $\frac{W}{8}\times 256$.
The encoder consists of three stacked bidirectional \ac{LSTM}-layers, each with 256 hidden nodes and a dropout rate of 0.5.

The Transformer decoder first applies a trainable embedding layer on the history of characters, and then adds sinusoidal absolute positional encoding to enable the attention modules to learn the order of characters.
Note, that no additional positional encoding is applied to the outputs of the encoder since it is learned autonomously.
Afterwards, to learn long range dependencies between the decoder features, a self-attention module, and to couple the textual with the visual features, a mutual-attention module are applied, each with 8 heads and 512 nodes.
Last, a \ac{PFF} layer with an output dimension of 256 is applied.
Two blocks each consisting of one self-attention, one mutual-attention, and one \ac{PFF} are stacked to form the final output.
Ultimately, a dense layer maps the features to the alphabet size $|A|+1$ (including the \ac{EOS}-token).

Each \ac{MHA} requires three inputs, a query $Q$, key $K$, and value $V$, whereby the output $Y$ is computed as
\[
    Y_i = \text{Softmax}\left(\frac{Q'_i\cdot K'}{\sqrt f}\right)\cdot V'\;,
\]
where $f$ is the number of features, $i$ is the $i$-th entry of each matrix, and $Q',K',V'$ are obtained by three independent linear transformations using the matrices $W_Q, W_K, W_V$.
Several so-called attention heads using different transformations are computed in parallel and concatenated afterwards.
In a self-attention module, all inputs are identical: $Q\equiv K\equiv V$.

The backbone and encoder form a sub-network that can be applied for \ac{HTR} autonomously if decoded with \ac{CTC} after appending an additional dense layer to map the encoder outputs to $|A|+1$ (including the \emph{blank} token).
These outputs, the \ac{CTC} confidences, are later used during training and decoding (see next Sections).

\subsection{Training}
\label{sec:training}

The total loss $L_\text{tot}$ for training comprises two components: the \ac{CTC}-loss $L_\text{CTC}$ computed on the CTC confidences, and the character-wise Cross-Entropy-loss $L_\text{CE}$ applied to the decoder outputs.
Both losses are weighted by the factor $\lambda_{CTC}\in\left[0,1\right]$ which denotes the influence of $L_\text{CTC}$:
\[
L_\text{tot} = \lambda_{CTC} \cdot L_\text{CTC} + (1 - \lambda_{CTC}) \cdot L_\text{CE}
\]
Throughout this paper, we set $\lambda_{CTC}=0.3$ which is a compromise of having a greater influence of the decoder but also a non-negligible encoder contribution.

Teacher forcing is applied to speed-up training by presenting the \ac{GT}-sequence, prefixed with a \ac{SOS} token, to the decoder during training (see Figure~\ref{fig:network-architecture}).
The characters in the ``future'' are masked so that the network has no access for predicting the ``next'' character.
Note that we do not apply label smoothing.

Our setup computes the exponential mean average with a decay of 0.99 of all weights during training to obtain the final model weights for evaluation.
We use an ADAM-optimiser with global gradient clipping of 5.0.
The learning rate is increased linearly from 0.0 to a maximum of 0.001 after five epochs and then reduced by a factor $\sqrt{5/i_\text{epoch}}$, where $i_\text{epoch}$ is the current epoch.
Training is finished if the model saturates on the validation dataset.
The best model on the validation dataset is stored as final model.

\subsection{Inference}

\begin{figure}[tbp]
    \centering
    \includegraphics[width=0.8\linewidth]{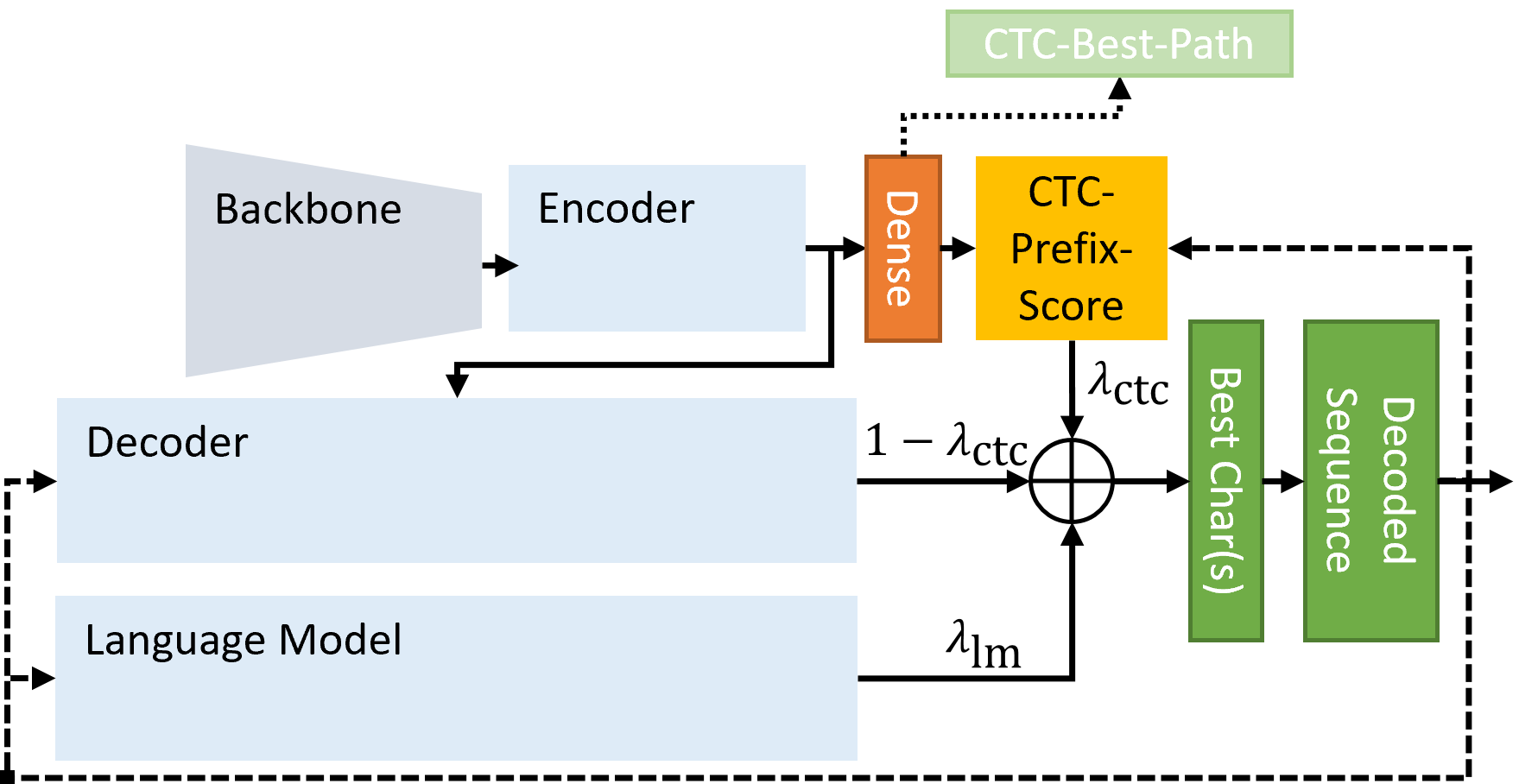}
    \caption{The decoding scheme using beam search. The character costs of the \ac{CTC}-Prefix-Score, the \ac{S2S} decoder, and \ac{LM} are weighted ($\lambda_\text{ctc}, \lambda_\text{lm}$) and summed up to obtain the next best characters for decoding.
    The history of the decoded sequence (dashed arrows) is required in all three modules to predict the next character.
    For comparison, best path decoding can be applied directly on the \ac{CTC} confidences.}
    \label{fig:decoding}
\end{figure}

During inference (see Figure~\ref{fig:decoding}), the decoder transcribes the input character by character starting with a \ac{SOS} token using beam search with a beam width of $n_\text{beams}$.
The confidences for the next character are a combination of three parts that each contributes costs $C\in\mathbb{R}^{\left|A\right| + 1}$ (alphabet size plus \ac{EOS}), the negative logarithms of the confidences:
\begin{itemize}
    \item The \textbf{decoder} directly outputs confidences for the next character $C_\text{CE}$. 
    \item Based on the \ac{CTC} confidences which are computed by the additional dense layer after the \textbf{encoder} (orange), the \ac{CTC}-Prefix-Score $C_\text{CTC}$ of the next sequence is computed (see next Section).
    \item A \ac{LM} which computes confidences for the next character $C_\text{LM}$ solely based on the previous sequence (see Section~\ref{sec:method:lm}).
\end{itemize}

The total costs $C_\text{tot}$ are computed as
\[
C_\text{tot} = \lambda_{CTC} \cdot C_\text{CTC} + (1 - \lambda_{CTC}) \cdot C_\text{CE} + \lambda_{LM} \cdot C_\text{LM}\;
\]
where $\lambda_\text{LM}\in\left[0,1\right]$ is the weight of the \ac{LM}.

\subsubsection{CTC Prefix Score}

The CTC-Prefix-Score decoding of Watanabe et al.~\cite{watanabe2017hybrid} is an addition to the \ac{S2S} decoding scheme by penalising paths in the beam search that contradict the \ac{CTC} confidences.
Here, we provide only a coarse description of their algorithm and refer to the original paper for the algorithms and the mathematical details.
For a deeper description of the \ac{CTC} algorithm we refer to the original paper of Graves et al. \cite{graves_connectionist_2006}.

The forward variables of \ac{CTC} allow to compute the probability of a sequence of characters by summing up the confidences of all possible paths that result in this sequence.
The \ac{CTC}-Prefix-Score is the total score of all paths that start with a given but end with an arbitrary sequence.
Watanabe et al.~propose to add this score to rescale each path during beam search decoding of \ac{S2S}.
If a path emits an \ac{EOS}-token the full \ac{CTC}-score is used for rescaling.
This overall setup prevents paths that prematurely end or skip characters since those receive a bad \ac{CTC}-Prefix-Score.
Unfortunately, the computation is slow since for each beam all possible path endings must be summed up for each decoding step.

\subsubsection{Beam Search}

The applied beam search tracks the best $n_\text{beams}$ character sequences that have not yet finished, and all (or only the best) completed sequences, i.e., those that predicted an \ac{EOS}-token.
By default, for each unfinished sequence, $C_\text{tot}$ and its parts are computed.
Hereby, as already stated, the computation of the \ac{CTC}-Prefix-Score is costly if evaluated for every possible next character.
To speed up the computation, we determine so-called ``pre-beams'' (see, e.g., the usage in \cite{li2020espnet}) which splits the computation of $C_\text{tot}$ in 
\[
C_\text{pre} = (1 - \lambda_{CTC}) \cdot C_\text{CE} + \lambda_{LM} \cdot C_\text{LM}
\]
and 
\[
C_\text{tot} = \lambda_{CTC} \cdot C_\text{CTC} + C_\text{pre}\;.
\]
After $C_\text{pre}$ is computed, the search space is narrowed down by reducing the possible beams to a maximum of $1.5\cdot n_\text{beams}$.
Only the \ac{CTC}-Prefix-Scores for the remaining characters $C_\text{CE}\in\mathbb{R}^{1.5\cdot n_\text{beams} + 1}$ are evaluated.
Then the actual $n_\text{beams}$ best beams are selected.
This simplification (if $1.5\cdot n_\text{beams} < |A|$) assumes that the best characters of the decoder are similar to those of the encoder and will lead to only minor differences in the result.
To the best of our knowledge, there is no open-source Tensorflow implementation for beam search supporting pre-beams which is why we share our code on \href{https://github.com/Planet-AI-GmbH/tfaip-hybrid-ctc-s2s}{GitHub}.

\subsection{Language Model}
\label{sec:method:lm}

Our character-based \ac{LM} is a traditional Transformer (equivalent to the decoder of  Figure~\ref{fig:network-architecture} without mutual-attention) which stacks six self-attention and subsequent \ac{PFF} layers with 512 output nodes, and 2,048 nodes in the first \ac{PFF} dense layer.
We use 8 attention heads and do not apply dropout.
The embedding layer has a dimension of 128.
The training hyper-parameters are identical to the ones of training the \ac{HTR} model.

Text samples are transformed and augmented by applying random transformations on the text:
\begin{enumerate}
    \item The text is randomly cropped (start and end) to a length between 768 and 1,024 characters.
    \item If the first character is upper case, its case is switched with a probability of 0.5, while if it is lower, it is switched with a probability of 0.1. 
    \item The IAM dataset does not follow traditional spacing rules, e.g., by inserting spaces before punctuation.
    Therefore, we add an additional rule for IAM only that randomly inserts spaces before, after, or on both sides of \texttt{;,-:'"?!.[]()\{\}} with a probability of 0.1.
    This results in a \ac{LM} that will have uncertainty on the exact location of spaces and special characters, but high ones on characters.
\end{enumerate}

\subsection{Synthetic Line Generation}

\begin{figure}[tbp]
    \centering
    \includegraphics[width=\linewidth]{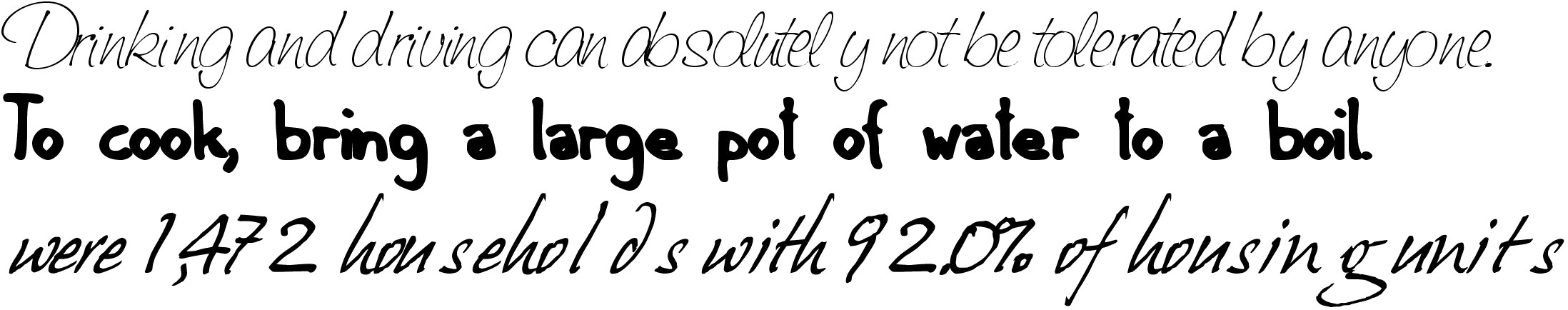}
    \caption{Three synthetically rendered lines based on the English text corpus.}
    \label{fig:synth-lines}
\end{figure}
Synthetic handwritten lines are used to pre-train the visual features and also to train the ``intrinsic'' \ac{LM} of the decoder.
To render synthetic lines (see Figure~\ref{fig:synth-lines} for some examples), we collected several cursive computer fonts.
The sentences are generated analogously to the cropping rules used for the \ac{LM} (see Section~\ref{sec:method:lm}).
Data augmentation during preprocessing (see Section~\ref{sec:preproc}) then varies the renderings with clean synthetic computer fonts to make the network more robust against real-world data.

\section{Results}

In this section, we present our evaluation results.
The network architecture and decoding are implemented in the open-source framework \emph{tfaip} \cite{wick2021tfaip} which is based on Tensorflow.
The encoder plus backbone, decoder, and \ac{LM} comprise 3.2, 1.6, and 19.0 million parameters, respectively.
The combined model has about 24 million parameters in total.

First, we show the accuracies of our trained \acp{LM} and the performances of the pretrained models applied to the three datasets.
Then, we measure the \acp{CER} when training on real data, optionally starting from the pretrained models.
Next, the \ac{LM} is added and  $n_\text{beams}$ is varied.
We conclude our experiments with an ablation study and a comparison to related work.

\subsection{Language Models}

\begin{table}[tbp]
    \centering
    \caption{Top-1 and top-10 accuracies of the trained \acp{LM} for correctly predicting the next character on the listed dataset.
    The model was trained on an additional corpus for the respective language.
    }
    \label{tab:result:language-model}
    \begin{tabular}{ll|rr}
    \toprule
        \textbf{Dataset} & \textbf{Language} & \textbf{Top-1} & \textbf{Top-10} \\
    \midrule
        IAM & en & 58.1\% & 90.4\% \\
        StAZH & de-ch & 52.5\% & 87.3\% \\
        Rimes & fr & 58.5\% & 92.8\% \\
    \bottomrule
    \end{tabular}
\end{table}

Table~\ref{tab:result:language-model} shows the top-1 and top-10 accuracies, to correctly predict the next character, of the three \acp{LM}.
The accuracies are stated for the validation datasets of IAM, StAZH, and Rimes, including the augmentations applied in Section~\ref{sec:method:lm}.
The independent training data was chosen from the respective datasets (see Section~\ref{sec:data}).

All \acp{LM} yield a top-1 accuracy of over 50\% and a top-10 accuracy of about 90\%.
Note that, as expected, the \ac{LM} for StAZH is the worst because the data of the training corpus is contemporary Swiss-German (de-ch) while StAZH comprises historic documents where spellings are different.
Since the French (fr) corpus is larger than the English (en) one, the accuracy is higher.
Other reasons for the lower accuracy on English are the different whitespace rules and the additional augmentation to map this.

\subsection{Pretraining on Synthetic Data}
\label{sec:res:synth}

\begin{table}[tbp]
    \centering
    \caption{\ac{CER} given in percent of the models pretrained on artificial lines synthesised for the respective language.
    We list the \ac{CER} for using solely the encoder with \ac{CTC} best path decoding and our proposed \ac{CTC}/Transformer combination.}
    \label{tab:pretraining}
    \begin{tabular}{c|cc|cc}
    \toprule
        & \multicolumn{2}{c|}{\bf \ac{CTC}} & \multicolumn{2}{c}{\bf \ac{CTC}/Trafo} \\
       \bf Dataset & \bf Val. & \bf Test & \bf Val. & \bf Test \\
    \midrule
        IAM & 17.4 & 19.5 & 15.3 & 17.3\\
        StAZH & 64.7 & 65.1 & 64.6 & 64.3 \\
        Rimes & 24.3 & 25.1 & 22.4 & 23.0 \\
    \bottomrule
    \end{tabular}
\end{table}

Table \ref{tab:pretraining} lists the \acp{CER} for the pretrained models which are trained exclusively on synthetic data generated for each of the three languages.
Naturally, since only synthetic data was used, the models perform very poorly on real world data, especially on StAZH which is the most difficult dataset.
Combining \ac{CTC} and the Transformer for decoding yields slightly better results than the standalone encoder with \ac{CTC} which is explained by the fact that the decoder learned an intrinsic \ac{LM}.

\subsection{Training on Real Data}

\begin{table}[tbp]
\centering
\caption{Influence of using pretrained models. The \ac{CER} is given in percent.
We list the \ac{CER} for using solely the encoder with \ac{CTC} best path decoding and our proposed \ac{CTC}/Transformer combination.
}
\label{tab:real-data}
    \tabcolsep=0.21cm
\begin{tabular}{l|c|cc|cc}
\toprule
      &            & \multicolumn{2}{c|}{\bf Validation} & \multicolumn{2}{c}{\bf Test}  \\
      & \bf Pretr. & \bf CTC     & \bf CTC/Tr          & \bf CTC     & \bf CTC/Tr     \\
      \midrule
      
IAM   & No         & 3.63 & 3.38              & 5.47 & 5.10         \\
IAM   & Yes        & 3.17 & 2.60              & 4.99 & 3.96         \\
      \midrule
StAZH & No         & 3.22 & 2.93              & 3.05 & 2.81         \\
StAZH & Yes        & 3.17 & 2.64              & 3.06 & 2.66         \\
      \midrule
Rimes & No         & 4.84 & 4.47              & 4.31 & 3.88         \\
Rimes & Yes        & 4.87 & 4.14              & 4.25 & 3.49         \\
\bottomrule
\end{tabular}
\end{table}

Table \ref{tab:real-data} shows the \ac{CER} when training on the actual real-world data, optionally using the pretrained model from Section~\ref{sec:res:synth}.
On both the validation and test sets of the respective datasets, the \ac{CER} shrinks when using the pretrained model and CTC/Transformer for decoding.
However, the benefit is clearly higher for the IAM dataset.
If only using \ac{CTC} for decoding, pretraining has a smaller impact, and even leads to slightly worse results for StAZH on the test set and for Rimes on the validation set.
Since the \ac{CTC} branch only includes visual information, we explain this by the handwritten computer fonts which are more similar to IAM but differ considerably from the writings in the other two datasets.
This however shows that the pretrained decoder actually learned an intrinsic \ac{LM} because using the pretrained model always shows an improved \ac{CER} when using \ac{CTC}/Tr.

\subsection{Enabling the Language Model}

Even though the decoder of the \ac{HTR} model is already capable to learn a \ac{LM} as shown in the previous section, we examine if an additional specialised and larger \ac{LM} (see Section~\ref{sec:method:lm}) further improves the results.
The upper half of Table~\ref{tab:with-lm} summarises the results by setting $\lambda_\text{lm}$ to 0, 0.1, 0.5, and 1 ($n_\text{beams}=5$).
$\lambda_{lm}=0$ corresponds to using no \ac{LM} and thus to the values of Table~\ref{tab:real-data} (last column).
Here, we only show the results when combining all three decoding paths (\ac{CTC}/Tr/\ac{LM}), see Figure~\ref{fig:decoding}).

\begin{table}[tbp]
\centering
\caption{The upper half shows the influence of using an additional \ac{LM} during decoding by varying its weight $\lambda_\text{lm}$.
The bottom half investigates the influence of varying the number of beams $n_\text{beams}$ during beam search decoding.
All \acp{CER} and \acp{WER} are given in percent.}
\label{tab:with-lm}
\tabcolsep=0.21cm
\begin{tabular}{l|rrrr|rrrr}
\toprule
          & \multicolumn{4}{c|}{\bf CER [\%]}               & \multicolumn{4}{c}{\bf WER [\%]}            \\
          \bottomrule
          \toprule
\bf LM  & \bf 0       & \bf 0.1     & \bf 0.5     & \bf 1       & \bf 0        & \bf 0.1      & \bf 0.5      & \bf 1         \\
          \midrule
IAM       & 3.96 & 3.69 & 3.19 & 3.64 & 12.20 & 11.12 & 9.17  & 10.34  \\
StAZH     & 2.66 & 2.66 & 2.79 & 3.81 & 11.87 & 11.75 & 12.48 & 16.72  \\
Rimes     & 3.49 & 3.40 & 3.39 & 3.57 &  9.03  & 8.65  & 8.28  & 8.49   \\
\bottomrule
\toprule
\bf Beams     & \bf 1       & \bf 5       & \bf 10      & \bf 20      & \bf 1        & \bf 5        & \bf 10       & \bf 20        \\
          \midrule
IAM       & 5.81 & 3.19 & 3.17 & 3.13 & 14.09 & 9.17  & 9.00  & 8.93   \\
StAZH     & 4.37 & 2.79 & 2.79 & 2.79 & 15.54 & 12.48 & 12.48 & 12.50  \\
Rimes     & 4.10 & 3.39 & 3.19 & 3.19 & 9.19  & 8.28  & 7.61  & 7.61   \\
\bottomrule
\end{tabular}
\end{table}

On IAM, the \ac{LM} with $\lambda_\text{lm}=0.5$ yields clearly improved results ($\text{CER}=3.2\%$) which further reduces the \ac{CER} by 20\% (relative).
This shows that the intrinsic \ac{LM} of our \ac{HTR} model is not yet powerful enough to ``fully'' learn the English language.

In contrast, using the \ac{LM} trained on a contemporary Swiss-German corpus does not further improve the results on the historic StAZH dataset, instead the results worsen.
The reason is that the historic and contemporary writings are too different and therefore induce errors instead of correcting misspellings.

Similar to IAM, the French \ac{LM} fits well.
However, since the French alphabet size is larger than the English one, a higher beam count is required to obtain the best results as shown in the next Section.

\subsection{Varying the Beam Count}

The bottom half of Table~\ref{tab:with-lm} lists the \acp{CER} and \acp{WER} when varying $n_\text{beams}$ if using a \ac{LM} with $\lambda_\text{LM}=0.5$.
Setting $n_\text{beams}=1$ corresponds to best path decoding.
The results show that increasing $n_\text{beams}$ from one to five has the highest impact on the results.
On IAM, further increases only led to small improvements.
More beams on StAZH had no impact because the \ac{LM} does not fit well.
Setting $n_\text{beams}=10$ further reduced the error rates on Rimes which we explain by the bigger alphabet size due to characters with diacritics such as ``e,è,é,ê''.

\subsection{Ablation Study}

Furthermore, we performed an ablation study to measure the impact of pretraining, the inclusion of a \ac{LM}, and the decoder.
We chose IAM, set $\lambda_\text{LM}=0.5$ (if a \ac{LM} is used), and $n_\text{beams}=5$.
The results of the \acp{CER} and \acp{WER} are listed in Table~\ref{tab:ablation} (upper half).

\begin{table*}[tb]
    \centering
    \caption{
    Ablation study and comparison with related work. All \acp{CER} and \acp{WER} are given in percent.
    Transformers (Tr) in the encoder are comprised of stacked self-attention layers, Transformers in the decoder consist of mixed self- and mutual-attention layers.
    Our decoding is performed with 5 beams, exclusions with 20 beams are marked.
    ``+ Data'' denotes whether during training additional synthetic (syn) or real data is used, or if the IAM validation set is included (val).
    The LM column lists if a \ac{LM} (trained on external corpora) with open vocabulary (Open) or with a limited vocabulary size (50K works) was used during inference.
    The number of parameters (\#P) are given in millions.
    The last column lists the number of samples that can be processed per second using a batch size of 1 and a CPU only.
    }
    \label{tab:ablation}
    \tabcolsep=0.11cm
    \begin{tabular}{Nl|llll|rr|r|r}
    \toprule
       \multicolumn{1}{c}{} & \bf Authors & \bf Enc. & \bf Dec. & \bf + Data & \bf LM & \bf CER & \bf WER & \bf \#P & \bf \#/s \\
    \midrule
        \label{cmp:our:c-n-n} & Ours & LSTM & CTC & No & No & 5.47 & 17.93 & 3.2 & 10.77 \\
        \label{cmp:our:c-y-n} & Ours & LSTM & CTC & Syn & No & 4.99 & 16.85 & 3.2 & 11.57 \\
        \label{cmp:our:t-n-n} & Ours & LSTM & Tr  & No & No & 5.61 & 16.24 & 4.8 & 1.90 \\
        \label{cmp:our:t-n-y} & Ours & LSTM & Tr & No & Open & 14.38 & 18.25 & 24 & 0.50 \\
        \label{cmp:our:t-y-n} & Ours & LSTM & Tr  & Syn & No & 4.15 & 12.22 & 4.8 & 2.50\\
        \label{cmp:our:t-y-y} & Ours & LSTM & Tr  & Syn & Open & 6.46 & 13.38 & 24 & 0.86 \\
        \label{cmp:our:ct-n-n} & Ours & LSTM & CTC/Tr  & No & No & 5.09 & 15.88 & 4.8 & 0.69 \\
        \label{cmp:our:ct-n-y} & Ours & LSTM & CTC/Tr  & No & Open & 4.33 & 12.69 & 24 & 0.37 \\
        \label{cmp:our:ct-y-n} & Ours & LSTM & CTC/Tr  & Syn & No & 3.96 & 12.20 & 4.8 & 0.70\\
        \label{cmp:our:ct-y-y} & Ours & LSTM & CTC/Tr  & Syn & Open & 3.20 & 9.19 & 24 & 0.40 \\
        \label{cmp:our:ct20-y-y} & Ours (20) & LSTM & CTC/Tr & Syn & Open & \bf{3.13} & \bf{8.94} & 24 & 0.18 \\
    \midrule
        \label{cmp:ourval:ct-y-y} & Ours & LSTM & CTC/Tr  & Syn/Val & Open & 3.01 & 8.81 & 24 & 0.42 \\
        \label{cmp:ourval:ct20-y-y} & Ours (20) & LSTM & CTC/Tr & Syn/Val & Open & \bf{2.95} & \bf{8.66} & 24 & 0.18 \\
    \midrule
       \label{cmp:bluche} & Bluche \cite{bluche2017gated} & LSTM & CTC & No & 50K & 3.2\;\, & - & 0.75 & - \\
       \label{cmp:michael}  & Michael \cite{michael2019evaluating} & LSTM & S2S & Val & No & 4.87 & - & - & - \\
       \label{cmp:yousef} & Yousef \cite{Yousef2020AccurateDU} & FCN & CTC & No & No & 4.9\;\, & - & 3.4 & - \\
       \label{cmp:kang}  & Kang \cite{kang2020pay} & Tr & Tr & Syn & No & 4.67 & 15.45 & - & - \\
       \label{cmp:wick}  & Wick \cite{wick2021transformer} & Tr & Bi-Tr & No & No & 5.67 & - & - & - \\
      \label{cmp:diaz}   & Diaz \cite{diaz2021rethinking} & Tr & CTC & Syn/Real & Open & \bf{2.75} & - & $\approx12$ & -\\
      \label{cmp:li:base}  & Li \cite{li2021trocr} & Tr & Tr & Syn & No & 3.42 & - & 334 & - \\
      \label{cmp:li:large}  & Li \cite{li2021trocr} & Tr & Tr & Syn & No & 2.89 & - & 558 & - \\
    \bottomrule
    \end{tabular}
\end{table*}

As expected, the \ac{CER} always benefits from starting with a model pretrained on synthetic data.
In experiments \ref{cmp:our:c-n-n} and \ref{cmp:our:c-y-n} we only use the encoder combined with \ac{CTC}-best-path-decoding which therefore primarily shows the influence of pretraining on the visual backbone and the BiLSTM-encoder: using pretraining improves the \ac{CER} by about 9\%, relatively.
The Transformer decoder benefits by a larger margin (see, e.g., \ref{cmp:our:ct-n-n} and \ref{cmp:our:ct-y-n} where the \ac{CER} is reduced by about 23\%).
The reason is that the Transformer can learn an intrinsic \ac{LM} in addition to the improved visual features from the encoder.

Comparison of \ref{cmp:our:t-n-n}-\ref{cmp:our:t-y-y} to \ref{cmp:our:ct-n-n}-\ref{cmp:our:ct-y-y}, respectively, shows the impact of the \ac{CTC}-Prefix-Score during decoding which always leads to improved results, independent of using pretraining or a \ac{LM}.
Furthermore, it is astonishing that combining a \ac{LM} with a non-pretrained model results in very bad \acp{CER} (\ref{cmp:our:t-n-y}).
Visual inspection of the transcribed lines shows that there are many lines that are too short, i.e., words were skipped in between but mainly at the end.
Here, the \ac{CTC}-Prefix-Score (\ref{cmp:our:ct-n-y}) has a high impact since it penalises beams that do not cover all characters that are detected by the encoder.

Our best model (without additional training data, \ref{cmp:our:ct20-y-y}) is obtained by combining all proposed improvements (pretraining on synthetic data, a \ac{LM}, and adding the \ac{CTC}-Prefix-Score) and a bigger beam width of $n_\text{beams}=20$, resulting in a \ac{CER} of 3.13\% and a \ac{WER} of 8.94\%.

The last column of Table~\ref{tab:ablation} lists the number of lines that can be processed per second, using a batch size of 1 and CPU decoding only.
The numbers show that using \ac{S2S} instead of \ac{CTC} results in a slowdown of about 5-6.
The combination of \ac{S2S} with the \ac{CTC}-Prefix-Score further reduces the speed by a factor of about $3$.
Adding a \ac{LM} reduces the decoding speed by another factor of about $2$.
Thus, with $n_\text{beams} = 5$, this results in a speed reduction of about $26$.

Increasing $n_\text{beams}$ by a factor of 4 does reduce the decoding speed by another factor of 2 (only).
The dependency for low beam counts is non-linear since all beams can be processed in parallel.

\subsection{Comparison with the State of the Art}

Table~\ref{tab:ablation} also includes performances of previous publications.
Our best model yields a competitive \ac{CER} of 3.13\% which outperforms most previous models that, similar to our approach, do not include additional real-world data for training (\ref{cmp:bluche}-\ref{cmp:wick}).
We even outperform \ref{cmp:bluche} who use, in contrast to our open vocabulary \ac{LM}, a \ac{LM} with a limited vocabulary size of 50,000 words.
Comparing \ref{cmp:kang} to \ref{cmp:our:t-y-n}, which are comparable setups, in terms of including synthetic data and only using the Transformer for decoding, shows that our overall method is superior.
The reasons might be our preprocessing or the different network architecture.
\ref{cmp:wick} can be compared to \ref{cmp:our:t-n-n} since both use the same preprocessing pipeline but a different network architecture which explains the similar results.

Setup \ref{cmp:diaz} which achieves the current best value with a \ac{CER} of 2.75\% also includes a large amount of additional real-world data which is why a direct comparison must be considered as biased.
Nevertheless, this verifies our observations that including additional data (here the validation set of IAM) further improves the results up to a \ac{CER} of 3.01\% and 2.95\%, for $n_\text{beams}=5$ or 20 (\ref{cmp:ourval:ct-y-y} and \ref{cmp:ourval:ct20-y-y}), respectively.

The large model of the recent publication of Li et al. \cite{li2021trocr} (\ref{cmp:li:large}) also outperforms our best model.
We expect that the increased model size in cooperation with an improved text synthesising method, as used by \cite{li2021trocr}, explains their further improvements.
In contrast, their base model (\ref{cmp:li:base}) is significantly worse than our best model even though their model is significantly larger (334M vs 24M parameters).
While our model employs a \ac{LM}, Li et al. use a large decoder which should actually have more capacity for language modelling.
The benefit of our approach is that our \ac{LM} can be trained separately requiring only text data, which the transformer decoder as part of the full network has to be trained with fully annotated image and text data.

\section{Future Work}

In this paper, we proposed the combination of a CNN/LSTM-encoder and Trans\-former-decoder network for \ac{TLR}.
To solve intrinsic problems of the \ac{S2S}-approach, we added the \ac{CTC}-Prefix-Score \cite{watanabe2017hybrid} which was, to the best of our knowledge, not yet applied in the context of \ac{TLR}.
Furthermore, we added a separately trained Transformer as a \ac{LM}.
On the well-established IAM-dataset, we achieved a competitive \ac{CER} of 2.95\% significantly outperforming a current state-of-the-art model with ten times more parameters \cite{li2021trocr}.
Our approach is outperformed solely by a significantly larger model (\cite{li2021trocr}, 20 times more parameters), or by including an additional dataset with other real-world handwritten data \cite{diaz2021rethinking}.

A crucial disadvantage of our approach compared to \ac{CTC}-based ones is the very slow decoding time.
In practice, this must outweigh the gain in performance for the actual use-case.
Even though we already incorporated some methods to speed up decoding, e.g., by reducing the search space of the \ac{CTC}-Prefix-Score, other possibilities must be explored.
A straightforward approach is to use tokens instead of single characters.
Since a token comprises several characters, and can thus form syllables or even complete words, the number of iterated decoding steps is massively reduced.
For \ac{TLR}, this approach has already proven to be successful by \cite{li2021trocr}. \newline

\noindent\textbf{Acknowledgments}\;\;
This work was partially funded by the \acf{ESF} and the Ministry of Education, Science and Culture of Mecklenburg-Western Pomerania (Germany) within the project \acf{NEISS} under grant no ESF/14-BM-A55-0006/19.

\bibliographystyle{splncs04}
{\small \bibliography{references}}

\begin{acronym}[MDLSTM]
\acro{CER}{Character Error Rate}
\acro{CTC}{Connectionist Temporal Classification}
\acro{CNN}{Convolutional Neural Network}
\acro{ESF}{European Social Fund}
\acro{FCN}{Fully Convolutional Network}
\acro{GT}{Ground Truth}
\acro{HTR}{Handwritten Text Recognition}
\acro{LM}{Language Model}
\acro{LSTM}{Long-Short-Term-Memory-Cell}
\acro{MHA}{Multi-Head-Attention}
\acro{NEISS}{Neural Extraction of Information, Structure and Symmetry in Images}
\acro{OCR}{Optical Character Recognition}
\acro{PFF}{Pointwise Feed-Forward}
\acro{ReLU}{Rectified Linear Unit}
\acro{RNN}{Recurrent Neural Network}
\acro{S2S}{Sequence-To-Sequence}
\acro{TLR}{Text Line Recognition}
\acro{WER}{Word Error Rate}
\acro{EOS}[$\left<\mathrm{eos}\right>$]{End-of-Sequence}
\acro{SOS}[$\left<\mathrm{sos}\right>$]{Start-of-Sequence}
\end{acronym}


\end{document}